\documentclass[conference]{IEEEtran}
\usepackage{cite}
\usepackage{amsmath,amssymb,amsfonts}
\usepackage{graphicx}
\usepackage{textcomp}
\usepackage{xcolor}
\usepackage{subcaption}
\def\BibTeX{{\rm B\kern-.05em{\sc i\kern-.025em b}\kern-.08em
    T\kern-.1667em\lower.7ex\hbox{E}\kern-.125emX}}

\usepackage{hyperref}

\usepackage{algorithm}
\usepackage{algpseudocode}

\newcommand{\Input}[1]{\State \textbf{Input:} #1}
\newcommand{\Output}[1]{\State \textbf{Output:} #1}

\begin{document}

\title{FAST: Similarity-based Knowledge Transfer for Efficient Policy Learning}

\author{\IEEEauthorblockN{Alessandro Capurso*, Elia Piccoli*, Davide Bacciu}
\IEEEauthorblockA{\textit{Department of Computer Science} \\
\textit{University of Pisa, Italy}\\
\texttt{a.capurso1@studenti.unipi.it, elia.piccoli@phd.unipi.it}}
}

\maketitle

\begin{abstract}
Transfer Learning (TL) offers the potential to accelerate learning by transferring knowledge across tasks. However, it faces critical challenges such as negative transfer, domain adaptation and inefficiency in selecting solid source policies. These issues often represent critical problems in evolving domains, i.e. game development, where scenarios transform and agents must adapt. The continuous release of new agents is costly and inefficient. In this work we challenge the key issues in TL to improve knowledge transfer, agents performance across tasks and reduce computational costs. The proposed methodology, called FAST – Framework for Adaptive Similarity-based Transfer, leverages visual frames and textual descriptions to create a latent representation of tasks dynamics, that is exploited to estimate similarity between environments. The similarity scores guides our method in choosing candidate policies from which transfer abilities to simplify learning of novel tasks. Experimental results, over multiple racing tracks, demonstrate that FAST achieves competitive final performance compared to learning-from-scratch methods while requiring significantly less training steps. These findings highlight the potential of embedding-driven task similarity estimations.\\
\end{abstract}


\section{Introduction}
Learning is often thought of as a process rooted in interactions with the environment. Reinforcement Learning (RL) expands on this core concept by viewing learning as a trial-and error process, in which agents engage with the environment, make choices, and receive feedback in the form of reward or penalties. Traditionally, agents are trained from scratch to accomplish a single task, requiring extensive interactions with the environment to achieve proficiency far more than a human would need for comparable tasks. One primary challenge in RL is the substantial computational demands imposed by simulation, where training time and data requirements scale up for complex tasks.

In game development and other evolving environments it is expensive and sub-optimal to start at each iteration from zero. For example, during production at each iteration new features are added to the game or the scenario is modified. Given this setting, the following questions come as a logical consequence: what about agents deployed in the previous version? Is it possible to leverage their knowledge and simplify the training of new specialized agents?

In attempt to answer these questions, this work challenges the problem of Transfer learning (TL) \cite{tl} in RL. TL focuses on how to carry knowledge from one task, called \textit{source}, to another problem, called \textit{target}, where the main objectives are to reduce sample complexity, improve generalization, and speed up adaptation to new environments. However, this approach introduces some challenges as negative transfer and domain adaptation. Negative transfer occurs when knowledge from one domain negatively impacts performance into another. While TL seeks to enforce positive knowledge transfer, it can inadvertently introduce conflicting strategies or biases, impairing learning in specific tasks. This challenge is even more marked in scenarios where source and target task domains differ thus straightforward behavior adaptation is sub-optimal or worst case unsuccessful.

\begin{figure}[t]
\centerline{\includegraphics[scale=0.25]{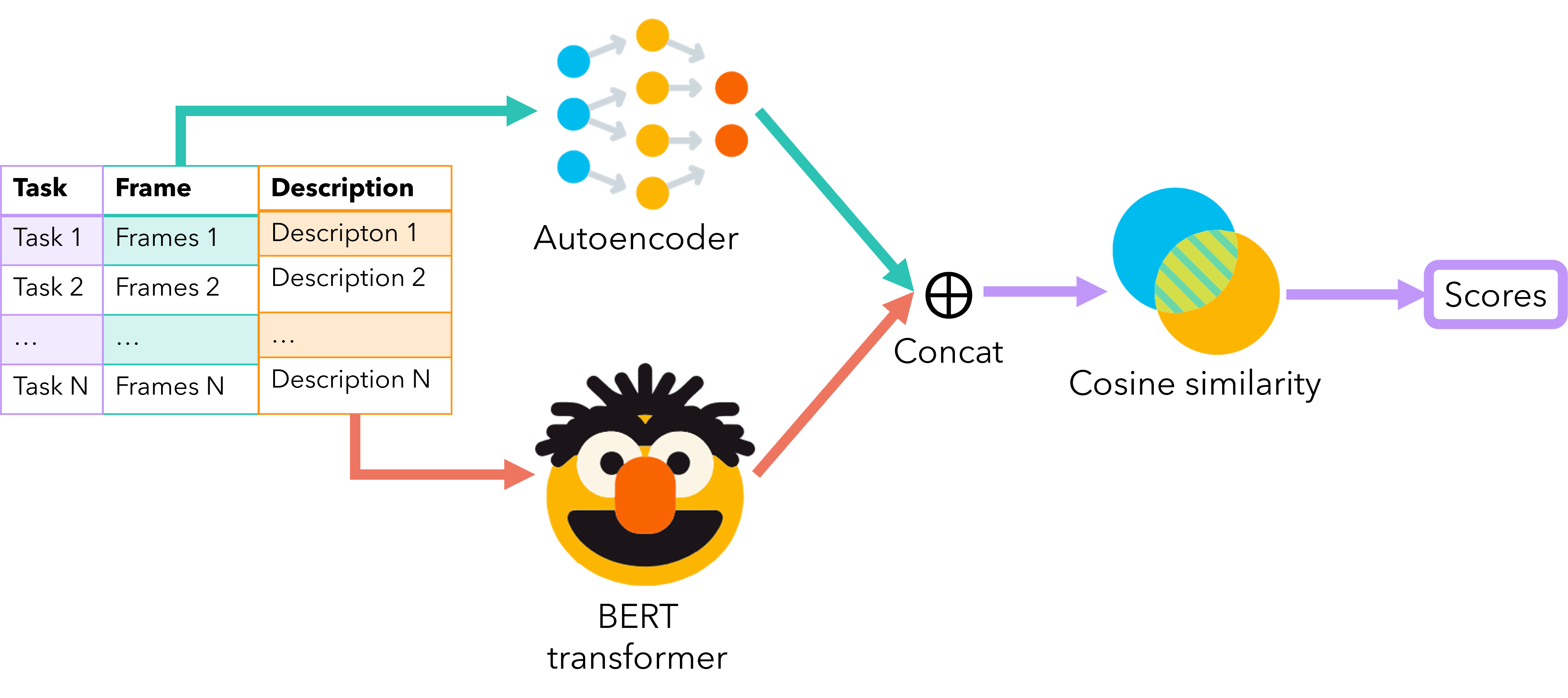}}
\caption{Schema of the designed similarity metric. Given a set of tasks, environment frames are processed using an autoencoder while goal descriptions are embedded leveraging BERT \cite{BERT}. The resulting representations are concatenated for each task and compared using cosine similarity, producing a vector of similarity scores.}
\label{fig:simschema}
\end{figure}

In order to overcome the problem of negative transfer, a solution is to exploit information about the \textbf{similarity} between different tasks. Given a set of source tasks, agents should be able to choose the one that is more similar to the target task, enforcing positive knowledge transfer and improving the training process. While the idea may seem simple, modeling and quantifying the similarity between different tasks and policies is still an open question \cite{tl}. Recalling the previous example about game development, this would allow to re-use between different iterations of previously trained agents and achieving a faster and adaptable deployment of new bots to newer versions. 

This work aims to address this challenge by proposing a combination of visual and textual information about tasks to compute a score that models the similarity between scenarios (\figurename \ref{fig:simschema}). Based on this information, learning agents can choose the policy of the most similar task from a repository of pre-trained policies. We test our solution on three different target tasks from the driving and racing scenarios in Highway-Env \cite{highwayenv}, achieving competitive performance against a tuned baseline, that learns the task from scratch, while using significantly less training steps. In the evaluation we do not consider only the reward but also other significant metrics that characterize the racing behavior of agents such as distance covered, number of laps completed and average speed.

\section{Background}
RL problems can be defined as Markov Decision Processes (MDPs), that are tuples  $\langle \mathcal{S}, \mathcal{A}, r, P, \gamma\rangle$, where $\mathcal{S}$ is the set of states, $\mathcal{A}$ is the set of possible actions, $r(s, a): \mathcal{S} \times \mathcal{A} \rightarrow \mathbb{R}$ is the reward function, $P: \mathcal{S} \times \mathcal{A} \times \mathcal{S} \rightarrow[0,1]$ is the transition function, $\gamma \in [0,1)$ is the discount factor. It is possible to define the return of an agent in an MDP as the sum of discounted rewards from a specific timestep $t$:
\begin{equation*}
    G_t = R_{t+1} + \gamma R_{t+2}+ \dots = \sum_{k=0}^{\inf} \gamma^k R_{t+k+1}
\end{equation*}
if $\gamma \thickapprox 0$ leads to a "myopic" evaluation, instead, if $\gamma \thickapprox 1$ leads to a "far-sighted" evaluation.

An RL agent aims to enhance the total reward it gets from the environment over time. The agent's objective is to learn a policy $\pi$, that maps states to actions to optimize the cumulative reward in any state. Formally, a policy $\pi$ is a distribution over actions $a$ given states s:
\begin{equation*}
    \pi (a|s) = P(A_t = a|S_t = s)
\end{equation*}
The policy depends only on the current state and it is stationary $A_t \sim \pi (\cdot | s), \forall t > 0$.\\

Deep Reinforcement Learning (DRL) combines the principles of RL with deep learning techniques. DRL methods can be categorized into three main approaches: value-based, policy-based, and actor-critic. In value-based methods, the goal is to estimate the optimal action-value function (Q-function) that predicts the future expected reward for taking a specific action in a given state, i.e. \texttt{DQN} \cite{dqn}. Policy-based methods, on the other hand, directly optimize the policy without estimating the value function, i.e. \texttt{PPO} \cite{ppo}. Actor-critic methods combine elements of both value-based and policy-based approaches, maintaining two sets of parameters: the \texttt{Critic} updates action-value function parameters $w$ and the \texttt{Actor} updates the policy parameters $\theta$. The critic estimates the value function, which acts as a reference for the actor’s gradient update. By offering more precise estimates of the expected return and decreasing the variance of the updates, it aids in stabilizing the learning process. A notable algorithm in this family, upon which is build our work, is \texttt{Soft Actor-Critic} (SAC) \cite{sac}. This method enhances the standard actor-critic by incorporating an entropy term into the objective function. The inclusion of entropy encourages exploration and adds stability during training. The maximum entropy objective can be expressed as:
\begin{equation*}
    J(\pi) = \sum_{t=0}^{T} \mathbb{E}_{(s_t, a_t) \sim \rho_\pi} [r_t + \alpha H(\pi(\cdot | s_t))] 
\end{equation*}
the $H(\pi(\cdot | s_t))$ represents the entropy of the policy $\pi$ at state $s_t$, and $\alpha$ is a temperature parameter that balances the reward and entropy terms. A higher $\alpha$ places more emphasis on exploration, while a lower $\alpha$ focuses more on maximizing the reward.

\section{Related works}
\subsection{Transfer Learning}
Over the years, various approaches have been proposed, implemented, and tested to improve task generalization and transfer learning (TL). One notable example is Teacher-student \cite{teacher_student}, which relies on two main components: an expert policy, \textit{teacher}, trained on specific tasks, and a unified learning policy \textit{student} that integrates and generalizes the acquired knowledge across different tasks. A teacher-student framework is presented in \cite{policy_dist} where a set of teacher policies share their knowledge with a single student policy. An evolution of this idea is Actor-Mimic \cite{actor_mimic} that employs imitation learning to replicate expert teacher policies for specific tasks. This helps students learn more quickly by leading them through task-specific guidelines, which can speed up training and enhance overall understanding. A similar scenario has been studied in Continual Learning (CL) \cite{continual_learning} where RL agents learn multiple tasks sequentially, preserving prior knowledge while acquiring new skills. Notable examples of this research direction are \cite{progress_compress}-\cite{soft_mod} where agents can extend their knowledge by adding new skills while they maintain a clear separation between different tasks in order to avoid catastrophic forgetting and loss in performance. Meta Learning  \cite{meta_learning} follows the idea of fast adaptation providing multiple solutions that tackle the aforementioned problems. In \cite{MAML}-\cite{VariBAD} agents can easily adjust to new scenarios based on their characteristics by either updating a set of shared parameters or adapting their policy. On the other hand, \cite{zero_shot_ll} propose a strategy that leverages high-level task descriptors for zero-shot knowledge transfer, helping the model generalize to new tasks without requiring additional training data.

\subsection{Task Similarity in Transfer Learning}
Task similarity in TL refers to estimating the degree of resemblance between the source task and the target task. When the source and target tasks are highly similar, the model can leverage learned representations and patterns more effectively, reducing the need for extensive retraining and improving generalization. Conversely, low task similarity may lead to negative transfer, where the knowledge from the source task interferes with learning the target task, degrading performance. Identifying and quantifying task similarity is crucial for selecting appropriate source tasks and designing effective transfer learning strategies.
Similarity can be defined in terms of shared input features, output labels, underlying data distributions, or structural properties of the tasks. Some methods \cite{spirl, simpl, sbmbrl} learn latent representations, which represent the skills’ embeddings or dynamics, that are later exploited to plan in the skill space or to easily adapt to new tasks via similarity between representations. Similarly, the concept of option can be used to decompose the problem and estimate affinity among problems \cite{subgoal_discovery} or to guide the learning by defining a library of behaviors that can be used to solve the task based on their effectiveness \cite{do_as_i_say}. Given a set of source tasks, some methods learn to estimate the similarity by predicting its value to choose the best one \cite{inter_task_transferability}. On the other hand, \cite{IKH}-\cite{polsubspace} leverage all the source tasks information via a linear combination to effectively solve the target task.
The similarity of the tasks can be also estimated extracting and leveraging the visual and textual features of the video of the environment to perform zero-shot learning \cite{BehAVE}.

\section{Framework for Adaptive Similarity-based Transfer (FAST)}
The framework proposed in this work aims to challenge several problems that have been analyzed in the previous sections. In particular, the main focus is to define a metric to quantify similarity between tasks, reduce computational overhead by leveraging a more detailed representation of task relationships, enabling more adaptable and effective TL solutions while avoiding negative transfer. The method focuses on RL agents designed to exploit a knowledge base, that consists of a set of pre-trained policies over multiple source tasks, in order to solve a problem in a different environment. Throughout the training loop, agents can periodically compare the current task information with the one stored in the repository of prior tasks, if any of the source policies is evaluated to be similar then it is exploited for a fixed number of steps as supervision to enable knowledge transfer. The framework pipeline is built upon two complementary elements: (\textit{i}) the design of a similarity metric between tasks; (\textit{ii}) how to integrate such metric inside the RL training process.

\begin{figure}[t]
\centerline{\includegraphics[scale=0.1]{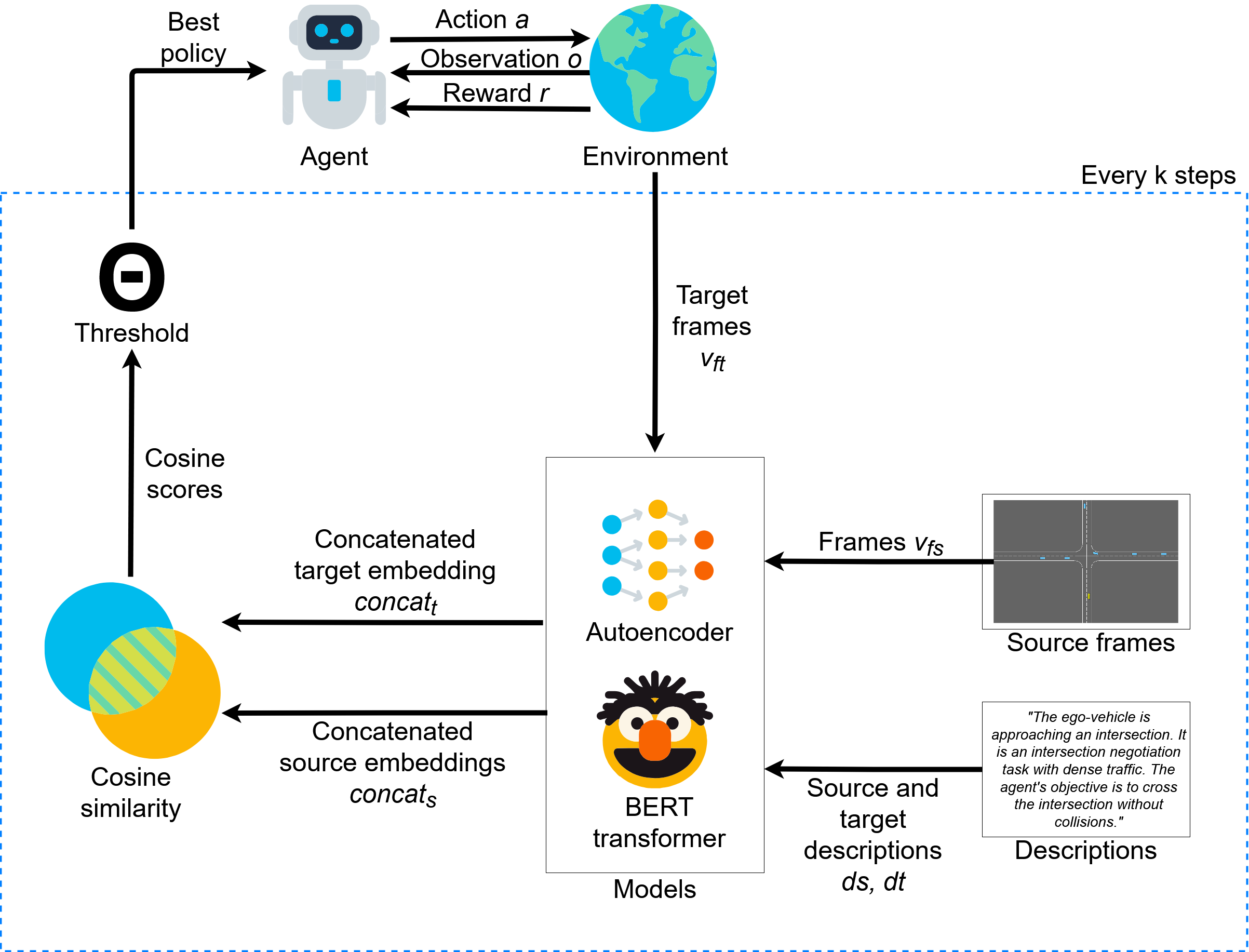}}
\caption{FAST schema. The framework integrates the designed similarity metric into RL training loop. Every \texttt{K} timesteps, frames and textual descriptions are extracted and embedded for both source and target task to estimate similarity. Scores are filtered based on $\Theta$, and the policy with the highest score is used for training over the next steps.}
\label{framework_schema}
\end{figure}

In the following sections, we analyze and present in detail the different elements that define our proposed methodology, also reported in Fig. \ref{framework_schema}.

\subsection{Similarity metric definition}
Similarity can be defined based on various factors, such as the overlap in state-action mappings, shared strategies, or comparable reward structures. However, despite its importance, there is no universally accepted framework for measuring similarity between policies and agents. This raises the first question we want to address in this work: \textit{How can we define and quantify similarity between different policies and agents?}

The first idea is to define a similarity metric based on frames of the environment that can also contain information about the position of the agent, its orientation, or the objects in its neighborhood. Unfortunately, if the frames belong to different environments, a pixel-wise comparison would not help to detect the similarity. To avoid this problem, the features of different tasks need to be comparable, so we create a shared latent space by leveraging an autoencoder to obtain a common representation (Fig. \ref{fig:simschema}).\\
The second element we believe can provide additional information useful to improve the similarity estimation between task is a textual task description. We assume that each task, both source and target, is paired with a brief natural language description of the underlying objective. As shown in Figure \ref{fig:simschema} this information is embedded in a latent space using BERT \cite{BERT}. By aligning both the visual and textual embeddings, we aim to create a unified representation that captures both the environmental context and the semantic intent of the task. The combined representation should help to avoid ambiguities that might arise from visual similarity alone and allow agents to distinguish between tasks with similar visual features but different goals.\\
The complete similarity schema is shown in Figure \ref{fig:simschema}. Considering two tasks, of which a set of four randomly sampled consecutive frames and a textual description are given, the frames are embedded using the pre-trained autoencoder and the textual descriptions are transformed using BERT. Once both the latent representations have been collected for each task, we concatenate the two embeddings and compute a similarity score between the concatenated representations through the cosine similarity. The similarity evaluation schema is then placed inside the main pipeline of our framework inside the dotted box, Fig. \ref{framework_schema}. The comparison is periodically assessed and the similarity scores are compared to a predefined threshold $\Theta$ in order to choose the policy that will guide agents' training.

\subsection{RL training pipeline}
In this work we make a distinction between two classes of tasks: source task and target task. The target task represents the current learning scenario and it is defined as a tuple $<v_{f_T}, d_T, \pi_T>$ where $v_{f_T}$ are the frames capturing the visual context of the task, $d_T$ is the textual description that details the task and $\pi_T$ is the policy to be trained. In contrast, a source task is defined as a previously solved task and serves as a basis for knowledge transfer, it is represented by a tuple $<v_{f_S}, d_S, \pi_S>$ where $v_{f_S}$ are the environment frames, $d_S$ is the textual description and $\pi_S$ is the pre-trained policy tailored to that specific task.
The first approach uses the metric periodically during the training process to choose the source policy that will guide the learning of the current task. It is a solution to guide and help the training process, it leads to good performances also in cases in which there are no similar source task available but it still require a training process to learn the task. On the other hand, the second approach directly applies the policy once it is selected performing zero-shot learning. In this case the training is not required but the performances are strictly dependent with the availability of similar source tasks. In this study, we opted for the first approach since it gives more flexibility in knowledge transfer between source and target policy.

Figure \ref{framework_schema} depicts the comparison process that is run every \texttt{K} timestep as dotted blue box. Running a comparison frequently would be costly in terms of time and resources and often unnecessary. For example, in successive timesteps, the states are likely to be similar enough that the metric would repeatedly select the same source policy. If the interval between comparisons is sparse enough, the agent may use different source policies. To balance this factor, we added a hyperparameter, \texttt{K}, to the training process, representing the number of steps between each comparison. This parameter helps to manage the trade-off between the cost of applying the metric and the accuracy of the source policy selection. Of course, this hyperparameter is strongly dependent on the target task.
Algorithm \ref{alg:transfer_alg} describes the RL pipeline that leverages task similarity to improve policy adaptation in the target task. It begins by loading a set of source policies $\pi_S$ and freezing them to prevent further updates. A target policy $\pi_T$ is initialized as the current policy $\pi_c$. The agent then interacts with the environment using the current policy $\pi_c$ to collect data and the target policy $\pi_T$ is updated using Soft Actor-Critic (SAC). This is a critical step of our methodology. The agent learns a specific policy for the target task $\pi_T$, which is the one being updated, while the policy to collect data and interact with the environment is $\pi_c$, which can also be the source policy. The interaction is depicted in the top part of Fig. \ref{framework_schema}. Every \texttt{K} steps, the algorithm collects frames from the current task and then evaluates task similarity. If the similarity exceeds a predefined threshold $\theta_s$, the current policy $\pi_c$ is updated by selecting the most suitable policy that could be either a source policy or the target one using Algorithm \ref{alg:select_best_policy_func}. This process continues until the specified number of timesteps \(n\) is reached. In our setup we aim at reducing the number of learning steps required while achieving competitive performance with baseline methods.

\begin{algorithm}[t]
    \caption{Pipeline pseudocode}\label{alg:transfer_alg}
    \begin{algorithmic}[1]
        \Input{source policies $\pi_{S}$, autoencoder $f_{ae}$, transformer $f_{t}$, n\_timesteps $n$, frequency \texttt{K}, similarity threshold $\theta$, target policy $\pi_{T}$, target task description $d_{T}$}
        \State Load and freeze source policies $\pi_S$
        \State count $= 0;$
        \State current\_policy $\pi_c = \pi_{T}$
        \For {$i=0 \rightarrow num\_timesteps $}
            \State Interact with environment and collect data using $\pi_c$
            \If{\textit{update condition}}
                \State update $\pi_T$ according using SAC
            \EndIf
            \State $count = count + 1$
            \If {$count \geq \texttt{K}$}
                \State Collect frames from current task using $\pi_{c}$\label{alg:get_frames_row}
                \State \textit{\# Update current policy via task similarity}
                \State $\pi_c = update\_policy(\pi_{T}, \pi_{S})$ \Comment{Algorithm \ref{alg:select_best_policy_func}}
                \State $count = 0$
            \EndIf
        \EndFor
	\end{algorithmic} 
\end{algorithm}

We now analyze more in detail Algorithm \ref{alg:select_best_policy_func} in order to understand step-by-step how using frames and text we are able to choose the optimal policy for the next training steps. As previously mentioned, this step is shown visually in Fig. \ref{framework_schema} inside the box. We assume that each source policy is equipped with a set of frames that characterize their behavior in the corresponding training environment, as well as a description of the task objective. When the procedure is triggered, a set of frames from the target task is collected $v_{f_T}$. This characterizes the states that the agent is currently visiting during its interaction with the environment. Frames from the source policies and target one are embedded using the pre-trained autoencoder $f_{ae}$:
\begin{equation*}
    emb_{f_S} = f_{ae}(v_{f_S}) \quad emb_{f_T} = f_{ae}(v_{f_T}).
\end{equation*}
The same transformation is applied to all task descriptions using the BERT transformer $f_t$:
\begin{equation*}
    emb_{d_t} = f_t(d_t) \quad emb_{dt} = f_t(d_t).
\end{equation*}
Once the embeddings are generated, a dedicated representation $\mathcal{R}$ of the target and source tasks are created by concatenating the two embeddings (frames and text) and normalized using MinMax scaling:
\begin{equation*}
    \mathcal{R_*} = min\_max\_scaling(emb_{f_*} \oplus emb_{d_*}).
\end{equation*}
we use * to indicate that the formula applies both to the target $T$ and source $S$ policies.

The target representation is then compared with the one of all the source tasks in our knowledge base using cosine similarity. From this comparison, a list of similarity scores is extracted. The scores are then compared with the threshold module indicated in Figure \ref{framework_schema} with the symbol $\Theta$. The threshold module is used to select only the tasks that have a certain degree of similarity, greater than $\theta$. If the threshold module returns no tasks, it means that at the current step no source policy is evaluated useful hence the target policy is used instead. Once the policy is selected, the agent uses the corresponding policy to act in the environment for the next \texttt{K} timesteps.

\begin{algorithm}[t]
    \caption{Evaluate similarity and choose policy} \label{alg:select_best_policy_func}
    \begin{algorithmic}[1] 
        \Input{target policy $\pi_{T}$, source policies $\pi_{S}$, target frames $v_{f_T}$, transformed and concatenated source data $concat_{s}$, transformed target description $emb_{dT}$, autoencoder $f_{ae}$, transformer $f_{t}$, similarity threshold $\theta$}
        \Output{best suited policy $\pi_{*}$}
        \State Compute target and source embeddings $emb_{f_*}, emb_{d_*}$
        \State Concatenate embedding to get final representation $\mathcal{R_*}$
        \For {$i=1 \rightarrow n\_source\_polcicy$}
            \State $cosine\_score = cos\_sim(\mathcal{R_{S_{i}}}, \mathcal{R_T})$
            \State $cosine\_list.append(cosine\_score)$
        \EndFor
        \State $idx, value = max(cosine\_list)$
        \If {$value \geq \theta$}
            \State \textit{\# Use the selected source policy}
            \State $\pi_{*} = \pi_{S}^{idx};$
        \Else
            \State \textit{\# Use the current target policy}
            \State $\pi_{*} = \pi_{T};$
        \EndIf
	\end{algorithmic} 
\end{algorithm}

\section{Experiments}
\subsection{Experimental Setup}
We use \texttt{stable-baselines3} (SB3) \cite{sb3} as a backbone, and build our framework\footnote{The code to replicate the experiments and agent's videos can be found on GitHub: \textit{https://github.com/Jek9884/FAST}.} as an extension on top of SB3's APIs. All the experiments are run using the Highway-env \cite{highwayenv} environment, that is compatible and accessible through the OpenAI Gymnasium Library \cite{gymnasium}. Highway-env offers traffic and racing track simulations that can be customized to test various driving strategies by adjusting traffic densities, behaviors, and dynamics. The agent engages with the surroundings by choosing actions like speeding up, slowing down, switching lanes, or staying in the same lane. The observation space consists of kinematic data, where each row in a matrix contains information describing the vehicles state, with the first row representing the controlled vehicle. The features considered throughout our experiments are the following:
\begin{itemize}
    \item \texttt{presence} is a bit representing if a vehicle is present, it is always 1 for the first row as it corresponds to the vehicle driven by the agent.
    \item \texttt{(x, y)} are the coordinates of the vehicle.
    \item \texttt{($v_x$, $v_y$)} are the velocity values of the vehicle on the x and y axis.
    \item \texttt{heading} indicates the heading of the vehicle, whose value is in radians.
    \item \texttt{long\_off} is the longitudinal offset of the vehicle on the closest lane.
    \item \texttt{lat\_off} is the lateral offset of the vehicle on the closest lane.
    \item \texttt{ang\_off} is the angular offset of the vehicle on the closest lane.
\end{itemize}
All feature values are standardized, and the vehicle coordinates are relative to the reference vehicle except for the reference vehicle itself, which is absolute. The maximum number of observed vehicles is 5, including the controlled vehicle; therefore, the final observation space is a matrix of dimensions 5 × 9 with the exception of the lane centering, which has the focus only for the controlled vehicle, hence the dimension is 1 × 9.

\begin{figure}[!bt]
    \centering
    \begin{subfigure}{0.13\textwidth}
        \includegraphics[width=\textwidth]{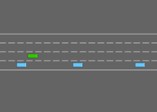}
        \caption{Highway}
        \label{fig:highway}
    \end{subfigure}
    \hfill
    \begin{subfigure}{0.13\textwidth}
        \includegraphics[width=\textwidth]{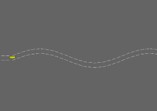}
        \caption{Lane centering}
        \label{fig:lane_centering}
    \end{subfigure}
    \hfill
    \begin{subfigure}{0.13\textwidth}
        \includegraphics[width=\textwidth]{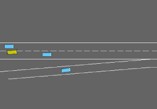}
        \caption{Merge}
        \label{fig:merge}
    \end{subfigure}
    \hfill
    \begin{subfigure}{0.13\textwidth}
        \includegraphics[width=\textwidth]{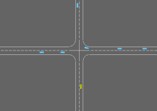}
        \caption{Intersection}
        \label{fig:intersection}
    \end{subfigure}
    \hfill
    \begin{subfigure}{0.11\textwidth}
        \includegraphics[width=\textwidth]{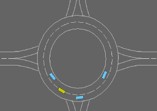}
        \caption{Roundabout}
        \label{fig:roundabout}
    \end{subfigure}
    \hfill
    \begin{subfigure}{0.13\textwidth}
        \includegraphics[width=\textwidth]{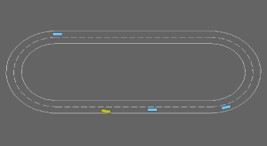}
        \caption{Indiana}
        \label{fig:indiana}
    \end{subfigure}
    \hfill
    \begin{subfigure}{0.20\textwidth}
        \includegraphics[width=\textwidth]{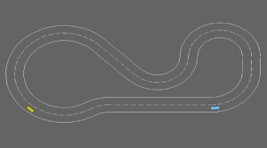}
        \caption{Racetrack}
        \label{fig:racetrack}
    \end{subfigure}
    \hfill
    \begin{subfigure}{0.20\textwidth}
        \includegraphics[width=\textwidth]{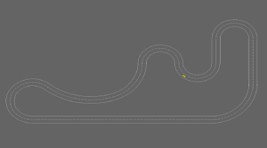}
        \caption{Custom racetrack}
        \label{fig:custom_racetrack}
    \end{subfigure}

    \caption{Highway-Env environment used in this work.}
    \label{fig:highway_tracks}
\end{figure}

We now provide a brief rundown of all the environments, both source and target tasks, that are used in our experimental evaluations. For each of them, we quickly describe the main objective and report visual representations of the scenarios in Figure \ref{fig:highway_tracks}. The environments used in this work, displaying different road scenarios, are the following ones:
\begin{itemize}
    \item \texttt{Highway} (\ref{fig:highway}), the controlled vehicle is on a straight road with many other vehicles that it must avoid.
    \item \texttt{Lane centering} (\ref{fig:lane_centering}), the vehicle must maintain the middle of a sinusoidal road with varying amplitude and frequency in each episode. The unique characteristic of this setting is the absence of other vehicles, requiring agents to concentrate solely on the controlled vehicle.
    \item \texttt{Merge} (\ref{fig:merge}), the vehicle is on a straight road and there is a road junction from the right, agent must avoid incoming cars.
    \item \texttt{Intersection} (\ref{fig:intersection}), the vehicle arrives at an intersection, and the goal is to safely cross the intersection.
    \item \texttt{Roundabout} (\ref{fig:roundabout}), the vehicle is at the entrance of a roundabout and should safely take one of the exits.
    \item \texttt{Indiana} (\ref{fig:indiana}), the vehicle should do an oval closed racetrack where the vehicle can loop till the end of the episode.
    \item \texttt{Racetrack} (\ref{fig:racetrack}), the vehicle is on a simple racetrack, made of many curves and straight lines.
    \item \texttt{Custom racetrack} (\ref{fig:custom_racetrack}), the vehicle is on a more complex racetrack. This circuit has been implemented by us, adding it to the pool of environments available, adding more complexity in fact it is significantly longer and it has different consecutive corners that makes the track more difficult to be completed.
\end{itemize}

As the proposed approach requires prior knowledge, the selected source policies are the following: \textit{Highway}, \textit{Indiana}, \textit{Intersection}, \textit{Lane Centering}, \textit{Merge}, and \textit{Roundabout}. To assess the performance of the pipeline as evaluation scenarios, three tracks, target tasks, of increasing difficulty were identified: \texttt{Indiana}, \texttt{Racetrack}, and \texttt{Custom Racetrack}. The first two circuits are existing tracks within the library, while Custom Racetrack was created specifically for these experiments. Although the Indiana circuit is among the source tasks, it was chosen as the starting point for experiments due to its simplicity. For the Indiana racetrack we excluded the pretrained Indiana policy because we wanted to test the performance of the transfer on a unknown task.

Preliminary tests revealed that especially on the Racetrack environments, the vehicle developed an unstable and suboptimal behavior, repeatedly moving one step forward and one step backward without making progress. This issue was primarily observed in complex track sections, i.e. consecutive or steep corners. Investigating on the default reward function, we noticed that it does not penalize backward movement:
\begin{equation*}
    r = a * \frac{v - v_{min}}{v_{max}-v_{min}} - b * collision
\end{equation*}
where $v$, $v_{min}$, $v_{max}$ are the current, minimum and maximum speed of the vehicle respectively, $a$ and $b$ are two coefficients and collision is $-1$ when a car accident happens or when the vehicle goes outside the track or 0 otherwise. To penalize the highlighted behavior, the final reward $r$ is multiplied by $-1$ if the car is not progressing through the track. An episode ends if either of the following condition is true: the car crashes into another vehicle, goes off the circuit lanes, or completes 150 seconds on the track without crashing or leaving the lanes. Relying only on the reward alone was not enough as it did not fully characterize the performance of the agents on the tracks. The main objectives while racing are the velocity, how much it can travel without going off-track or colliding with other vehicles and the numbers of laps completed. Hence, throughout our empirical evaluation we will report four different metrics averaged over multiple episodes on different seeds: average \texttt{reward} (AR), average \texttt{distance} (AD), average \texttt{number of laps} (AL) and average \texttt{speed} (AS).

All training experiments were conducted with a seed set to 42 to ensure replicability, while evaluation was obtained across multiple random seeds over 50 episodes. The vehicles are randomly placed at different track sections at each episode start, with vehicle counts randomized between 1 and 5. Having several elements that are randomized leads to very stochastic scenarios during evaluation. As a consequence, standard deviations of the reported metrics may have larger values than expected.

\subsection{Highway autoencoder}
The initial step involved training an autoencoder to extract the most relevant features from the frames of the Highway-env environments. To populate the training dataset, we run the policies from \cite{IKH} for a fixed amount of episodes while collecting frames from different task. The autoencoder network processes the input tensor with shape $600 \times 600 \times 4$, using three convolutional layers with 64, 128, 128 filters respectively, kernel size 5 and stride 3. The final tensor is then flattened and passed through a linear layer to produce an embedding of size \textbf{$128$}. The autoencoder was optimized to find the optimal set of hyperparameters using K-Fold.

\subsection{Results}
The first step was to create a dataset of frames for each of the source policies to be used in the transfer process. We ran 1000 episodes for each source policy and save the frames from episodes with the highest reward. The task descriptions are taken directly from the official documentation except for the Custom racetrack where we use the same description of the Racetrack since the objective is the same. As previously presented, the framework adds two new hyperparameters to the training phase of the agent, the frequency \texttt{K} and threshold $\theta$. To analyze in detail the contribution of the designed method, we divide the optimization process into two phases: (\textit{i}) we optimize the SAC hyperparameters for each target task; (\textit{ii}) we separately optimize the framework specific parameters.

The hyperparameters considered are: learning rate, batch size, total timesteps, $\gamma$ and $\tau$, SAC coefficient used for the soft update of the target critic. Any other parameter is kept with its default value. From the grid search, we obtained the best configuration indicated in Table \ref{tab:SAC_hp} for each of the target tasks. This hyperparameters are kept fixed for all the experiments. All the results of the proposed methodology are obtained using this predefined set of parameters without any additional and costly search. In our experimental evaluation we test the framework considering two possible variants: the first leverages only environment \textit{frames}, referred as (\texttt{F}), while the second exploits also the \textit{textual description} of the tasks, (\texttt{FT}). For both setups we fix the SAC hyperparameters and we only tune our method's parameters. Optimal values for the two scenarios are reported in Table \ref{tab:transfer_hp_of}.

\begin{table}[t]
    \caption{Hyperparameters values found using the grid search for all evaluation tracks.}
    \begin{center}
        \begin{tabular}{c|c|c|c}
            \textbf{Hyper}&\multicolumn{3}{|c}{\textbf{Best values found}} \\
            \cline{2-4} 
            \textbf{parameter} & \textbf{\textit{Indiana}}& \textbf{\textit{Racetrack}}& \textbf{\textit{Custom Racetrack}} \\
            \hline
            Learning rate & $1e-3$ & $1e-3$ & $1e-3$ \\
            Batch size & $1024$ & $1024$ & $1024$ \\
            Total timesteps & $1e7$ & $1e7$ & $1e7$ \\
            $\tau$ & $0.9$ & $0.5$ & $0.5$ \\
            $\gamma$ & $0.8$ & $0.99$ & $0.99$ \\
        \end{tabular}
        \label{tab:SAC_hp}
    \end{center}
\end{table}

\begin{table}[t]
    \caption{Hyperparameters values found using the grid search divided by track using only frames (F) or also textual information (FT).}
    \begin{center}
        \begin{tabular}{c|c|c|c}
            \textbf{Hyper}&\multicolumn{3}{|c}{\textbf{Best values found}} \\
            \cline{2-4} 
            \textbf{parameter} & \textbf{\textit{Indiana}}& \textbf{\textit{Racetrack}}& \textbf{\textit{Custom Racetrack}} \\
            \hline
            \texttt{F} - \texttt{K} & $10$ & $10$ & $100$ \\
            \texttt{F} - $\theta$ & $0.7$ & $0.5$ & $0.5$ \\
            \hline
            \texttt{FT} - \texttt{K} & $100$ & $1000$ & $1000$ \\
            \texttt{FT} - $\theta$ & $0.7$ & $0.6$ & $0.7$ \\
        \end{tabular}
        \label{tab:transfer_hp_of}
    \end{center}
\end{table}

\begin{figure}[!bt]
    \centering
    \begin{subfigure}{0.30\textwidth}
        \includegraphics[width=\textwidth]{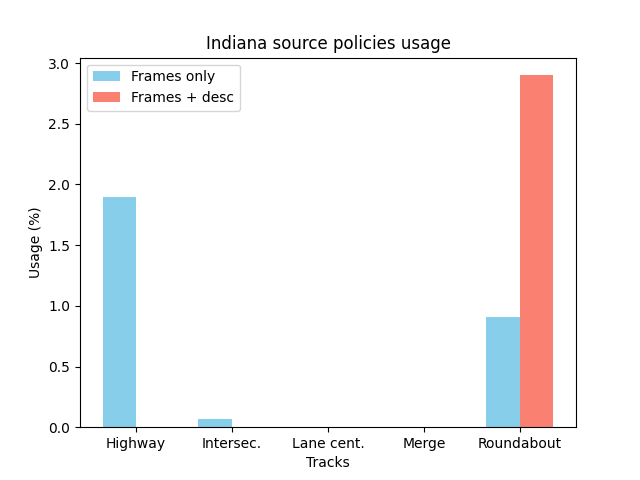}
        \caption{Indiana}
        \label{fig:indiana_transfer}
    \end{subfigure}
    \hfill
    \begin{subfigure}{0.30\textwidth}
        \includegraphics[width=\textwidth]{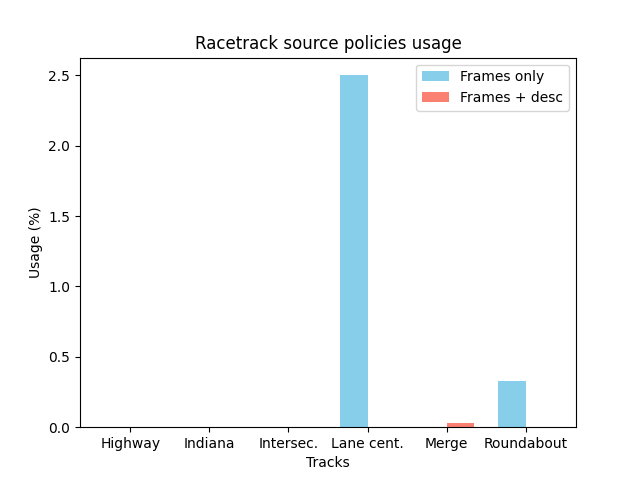}
        \caption{Racetrack}
        \label{fig:racetrack_transfer}
    \end{subfigure}
    \hfill
    \begin{subfigure}{0.30\textwidth}
        \includegraphics[width=\textwidth]{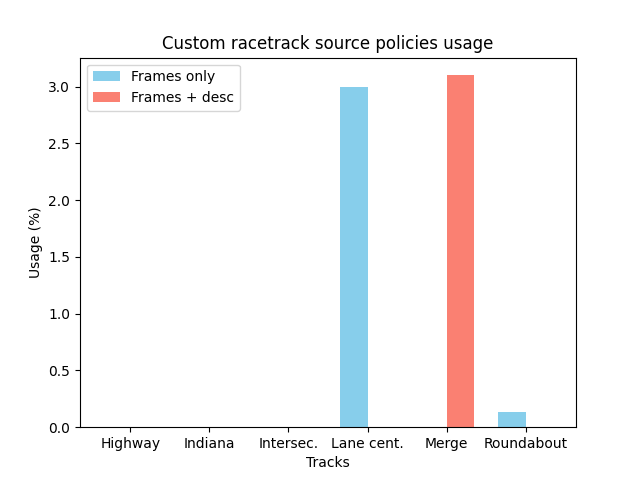}
        \caption{Custom racetrack}
        \label{fig:cr_transfer}
    \end{subfigure}
    \caption{Source policy usage in percent. The percentage is calculated by dividing the number of times each source policy is used by the total timesteps. The percentages do not add up to 100\%, as there are cases where no source policy was selected, and the target policy was used instead.}
    \label{fig:source_policy_percentage}
\end{figure}

Table \ref{tab:indiana_metrics_10_2_M} presents the performance evaluation on Indiana. The results show that the \texttt{FT} configuration achieves the highest average reward, nearly \textbf{double} that of the baseline and the frames-only approach. Similarly, the average distance covered is substantially greater, indicating improved exploration and performance thanks to a positive transfer from the source policies. The average number of laps (AL) showcase the same trend nearly doubling the values. The average speed (AS) remains relatively consistent across all approaches. These findings suggest that the addition of textual descriptions in this particular scenario enhances overall performance, leading to higher rewards and better task completion without compromising speed. Interestingly, in Figure \ref{fig:indiana_transfer} we report the usage of source policies for transfer learning. We can see that in this scenario, Highway and Roundabout are the predominant policies, which is coherent with the configuration of the target task compared to them, with the frame-text approach solely relying on the latter. We test the one-shot adaptability of the chosen source policies in the target task. Table \ref{tab:source_policies_on_racetrack} reveals that the source policies are very far from solving the task, nevertheless, they provide positive knowledge improving the performance leveraging our methodologies. 

Moving to more complex scenarios, Table \ref{tab:racetrack_metrics_10_2_M} presents the performance evaluation on the Racetrack environment across different models and training steps. At 10 million steps, the proposed frame-based model achieves the highest average reward, outperforming both the baseline and the frame-text model, it also demonstrates the highest average number of laps and average distance. At 2 million steps, the situation is overturned. The frame-text model shows strong performance, with an average reward superior than the baseline despite using only \textbf{one-fifth} of the training time. On the other hand, the frame-only model despite a significantly smaller reward shows competitive performance with the baseline. A similar trend of results can be observed in Table \ref{tab:custome_racetrack_metrics_10_2_M} for the Custom Racetrack. Both F and FT methods at 10M steps achieve better performance on average across several metrics, with frame only method taking the lead. The same standing is depicted while using 2M training steps with both methods outmatch the baseline with \textbf{5x} more training steps in all metrics except average reward. A key takeaway message of this evaluation is the reward function must be improved since it does not always correctly represent agents performance in the tracks. Interestingly, looking at Figure \ref{fig:racetrack_transfer}-\ref{fig:cr_transfer} our methods are consistent with the choice of source policies in the two tasks since the scenario share similar visual characteristics and textual description.

\begin{table}[t]
    \caption{Performance evaluation on Indiana. \texttt{B} is the SAC baseline, \texttt{F} the proposed approach with frames only and \texttt{FT} with frames and textual descriptions.}
    \begin{center}
        \begin{tabular}{c|c|c|c|c}
            & \textbf{AD} & \textbf{AR} & \textbf{AL} & \textbf{AS}\\
            \hline
            \texttt{B}  & $581_{\pm 916}$   &  $55.25_{\pm 89.59}$   & $1.89_{\pm 2.98}$  &  $36.46_{\pm 3.47}$ \\
            \texttt{F}  & $567_{\pm 692}$   &  $53.82_{\pm 67.82}$   & $1.87_{\pm 2.31}$  &  $36.33_{\pm 3.65}$ \\
            \texttt{FT} & $1119_{\pm 1958}$ &  $107.92_{\pm 191.83}$ & $3.71_{\pm 6.54}$  &  $36.35_{\pm 3.80}$ \\
        \end{tabular}
        \label{tab:indiana_metrics_10_2_M}
    \end{center}
\end{table}

\begin{table}[t]
    \caption{Performance evaluation on Racetrack using 2/10 million training steps. \texttt{B} is the SAC baseline, \texttt{F} the proposed approach with frames only and \texttt{FT} with frames and textual descriptions. Numbers indicates the million of steps performed during the training.}
    \begin{center}
        \begin{tabular}{c|c|c|c|c}
            & \textbf{AD} & \textbf{AR} & \textbf{AL} & \textbf{AS}\\
            \hline
            \texttt{B10}  & $818.64_{\pm 279.21}$  &  $207.53_{\pm 70.10}$  & $2.23_{\pm 0.78}$  &  $6.40_{\pm 1.10}$ \\
            \texttt{F10}  & $999.91_{\pm 336.73}$  &  $230.96_{\pm 68.28}$  & $2.75_{\pm 0.92}$  &  $7.34_{\pm 1.17}$ \\
            \texttt{FT10}  & $783.71_{\pm 228.82}$  &  $229.80_{\pm 59.95}$  & $2.14_{\pm 0.64}$  &  $5.74_{\pm 1.05}$ \\
            \hline
            \texttt{F2}   & $827.22_{\pm 436.88}$  &  $160.18_{\pm 88.26}$  & $2.27_{\pm 1.23}$  &  $8.48_{\pm 1.05}$ \\
            \texttt{FT2}   & $976.45_{\pm 298.62}$  &  $213.71_{\pm 63.67}$  & $2.63_{\pm 0.82}$  &  $7.34_{\pm 0.74}$ \\
        \end{tabular}
        \label{tab:racetrack_metrics_10_2_M}
    \end{center}
\end{table}

\begin{table}[h]
    \caption{Performance evaluation on Custom Racetrack using 2/10 million training steps. \texttt{B} is the SAC baseline, \texttt{F} the proposed approach with frames only and \texttt{FT} with frames and textual descriptions. Numbers indicates the million of steps performed during the training.}
    \begin{center}
        \begin{tabular}{c|c|c|c|c}
            & \textbf{AD} & \textbf{AR} & \textbf{AL} & \textbf{AS}\\
            \hline
            \texttt{B10}  & $345.05_{\pm 144.20}$  &  $209.27_{\pm 73.95}$   & $0.21_{\pm 0.22}$   &  $3.55_{\pm 3.63}$ \\
            \texttt{F10}  & $408.31_{\pm 127.16}$  &  $220.25_{\pm 42.88}$   & $0.27_{\pm 0.26}$   &  $3.17_{\pm 2.19}$ \\
            \texttt{FT10}  & $373.87_{\pm 175.95}$  &  $196.45_{\pm 84.56}$   & $0.25_{\pm 0.27}$   &  $4.61_{\pm 4.89}$ \\
            \hline
            \texttt{F2}   & $523.60_{\pm 320.70}$  &  $171.47_{\pm 123.90}$  & $0.44_{\pm 0.46}$   &  $5.97_{\pm 3.50}$ \\
            \texttt{FT2}   & $398.85_{\pm 266.68}$  &  $151.85_{\pm 118.21}$  & $0.32_{\pm 0.34}$   &  $4.10_{\pm 2.84}$ \\
        \end{tabular}
        \label{tab:custome_racetrack_metrics_10_2_M}
    \end{center}
\end{table}

\begin{table}[h]
    \caption{Performance of source policies on Indiana.}
    \begin{center}
        \begin{tabular}{c|c}
            \textbf{Policy} & \textbf{AR}\\
            \hline
            Random & $1.07_{\pm 2.27}$\\
            Highway & $3.36_{\pm 5.40}$\\
            Intersection & $13.64_{\pm 12.70}$\\
            Roundabout & $10.66_{\pm 6.71}$\\
        \end{tabular}
        \label{tab:source_policies_on_racetrack}
    \end{center}
\end{table}

\section{Conclusion}
In this work, we proposed FAST a RL framework to exploit previously learned tasks to train an agent on a new task preventing the negative transfer. The analysis showed that our proposed method can achieve competitive performance against an optimized baseline \textit{without any hyperparameter search}. The proposed training pipeline yields improvements in both learning efficiency and performance.

While textual descriptions initially appeared promising as a supplement to visual information, their impact was inconsistent, suggesting that textual data may only enhance learning in specific contexts. The lower transfer effectiveness of descriptions may be due to their quality; more precise and detailed descriptions may improve training performance. Nevertheless, FAST outperforms the vanilla approach, offering not only better results but also faster learning, exhibiting comparable outcomes at \textit{one-fifth} of the training cost. This efficiency advantage is valuable, as it may save time and resources allowing for faster prototyping and release of increasingly complex agents. In future works, we plan to refine and study more in detail textual descriptions, explore alternatives to cosine similarity, and integrate the framework in a continual scenario to improve performance in sequential learning.

\section*{Acknowledgment}
Research partly funded by PNRR- PE00000013 - "FAIR - Future Artificial Intelligence Research" - Spoke 1 "Human-centered AI", funded by the European Commission under the NextGeneration EU programme, and by the Horizon EU project EMERGE (grant n. 101070918).

\appendices
\section{Source task dataset generation}
To apply the transfer learning process a dataset of source experiences was needed, so we ran 1000 episodes for each source policy and saved the top episode with the highest rewards. The task descriptions were not generated from scratch; they were taken directly from the official documentation except for the Custom racetrack where we use the same description of Racetrack because the objective is the same. The descriptions are:

\begin{itemize}

    \item \textbf{Highway}: "The ego-vehicle is driving on a multilane highway populated with other vehicles. The agent’s objective is to reach a high speed while avoiding collisions with neighbouring vehicles."

    \item \textbf{Lane centering}: "The ego-veichle is driving on a single lane road which is not straight. The agent's objective is to maintain the car on the roadway.",

    \item \textbf{Merge}: "The ego-vehicle starts on a main highway but soon approaches a road junction with incoming vehicles on the access ramp. The agent’s objective is now to maintain a high speed while making room for the vehicles so that they can safely merge in the traffic."

    \item \textbf{Intersection}: "The ego-veichle if approaching an intersection. It is an intersection negotiation task with dense traffic. The agent's objective is to cross the intersection without collisions."

    \item \textbf{Roundabout}: "The ego-vehicle if approaching a roundabout with flowing traffic. It will follow its planned route automatically, but has to handle lane changes and longitudinal control to pass the roundabout as fast as possible while avoiding collisions."

    \item \textbf{Indiana}: "The ego-veichle is driving on a circuit. It is an oval circuit. The objective is to complete the track as fast as possible without touching the edges. "

    \item \textbf{Racetrack}: "The ego-veichle is driving on a racetrack. The agent's objective is to follow the tracks while avoiding collisions with other vehicles."
\end{itemize}

\section{Autoencoder}
\subsection{Dataset}
Training the Autoencoder required a dataset specific to the Highway-env environment. Since no pre-built dataset was available, a new dataset was generated using the provided source policies. For each policy, $1000$ episodes were sampled, with the last four frames of each episode saved. To maintain consistency, a default frame size of $600\times600$ was set, with upscaling applied where necessary due to varying task render dimensions. The final dataset comprises $6000$ examples.
\subsection{Model architecture and model selection}
\begin{figure}[t]
    \centerline{\includegraphics[width=0.49\textwidth]{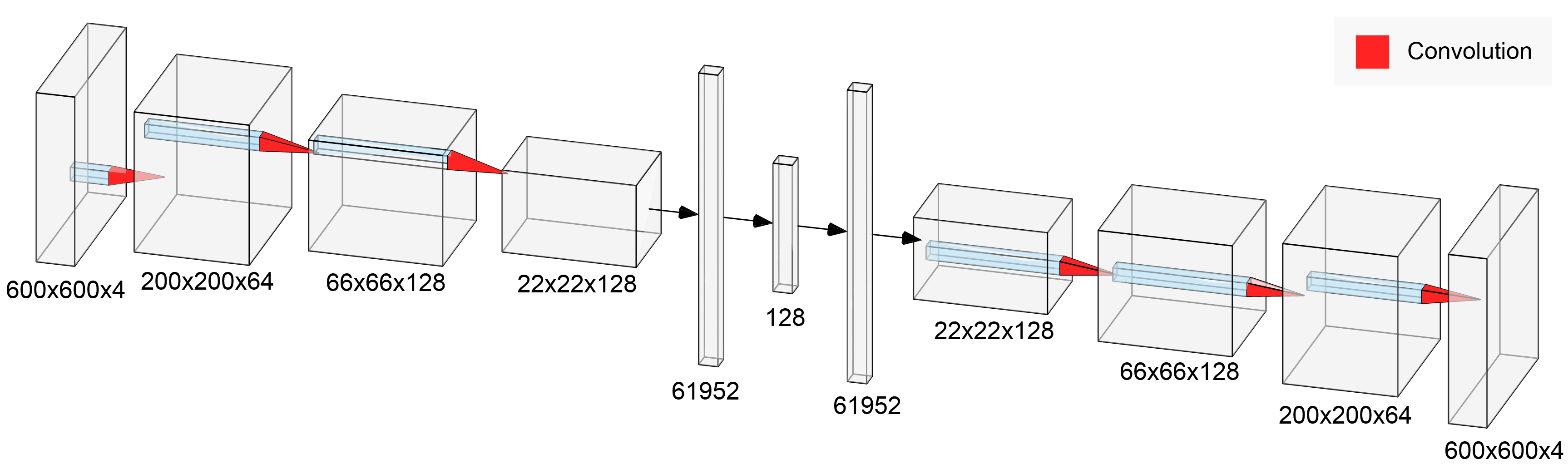}}
    \caption{Highway Autoencoder architecture. The encoder comprises five blocks, four of which are convolutional layers. After the final convolution, the output is flattened and fed into a linear layer, producing a latent representation vector of size $128$. The decoder is symmetrically structured, with five blocks mirroring the encoder.}
    \label{fig:highway_ae_arch}
\end{figure}

The network processes an input tensor of dimensions $600\times600\times4$. The three convolutional layers sequentially output tensors of sizes $200\times200\times64$, $66\times66\times128$, and $22\times22\times128$. This final tensor is then flattened into a vector of length 61952, which is passed through the linear layer to produce the embedding of size $128$ (see Figure \ref{fig:highway_ae_arch}).

To perform these transformations the convolution kernel size was set to $5$, the stride to $3$ and the padding to $1$.
Both weight decay and early stopping techniques were applied. All frames were converted to greyscale and normalized before being fed into the network. A grid search was conducted using MSE as the loss function to identify the optimal combination of hyperparameters, with the selected values presented in Table \ref{tab:HypGridHighway}.

\begin{table}[t]
    \caption{Values of hyper-parameters used in the Highway autoencoder grid search.}
    \begin{center}
        \begin{tabular}{c|c}
        \textbf{Parameter} & \textbf{Values}\\
                 \hline
        learning rate & 1e-3, 1e-5, 1e-7 \\[0.1cm]
        batch size & 32, 64, 128 \\[0.1cm]
        weight decay & 1e-3, 1e-5, 1e-7 \\[0.1cm]
        epochs & 1000 \\[0.1cm]
        latent dimension & 128 \\[0.1cm]
        eps & 1e-8 \\[0.1cm]
        patience & 5 \\[0.1cm]
        divergence threshold & 1e-5 \\[0.1cm]
        \end{tabular}
        \label{tab:HypGridHighway}
    \end{center}
\end{table}

The validation technique used to select and assess the model was the Holdout Validation with a split of 60\%-20\%-20\% of the dataset for the training, validation, and test set respectively. The metric used for the model selection was the MSE.

\subsection{Best model}
The optimal model identified had a learning rate of $1e-5$, a batch size of $128$ and a weight decay of $1e-7$. This model achieved a training score of $2.0383e-3$, a validation score of $2.0347e-3$ and a test score of $2.0365e-3$.

\begin{figure}[t]
    \centerline{\includegraphics[width=0.49\textwidth]{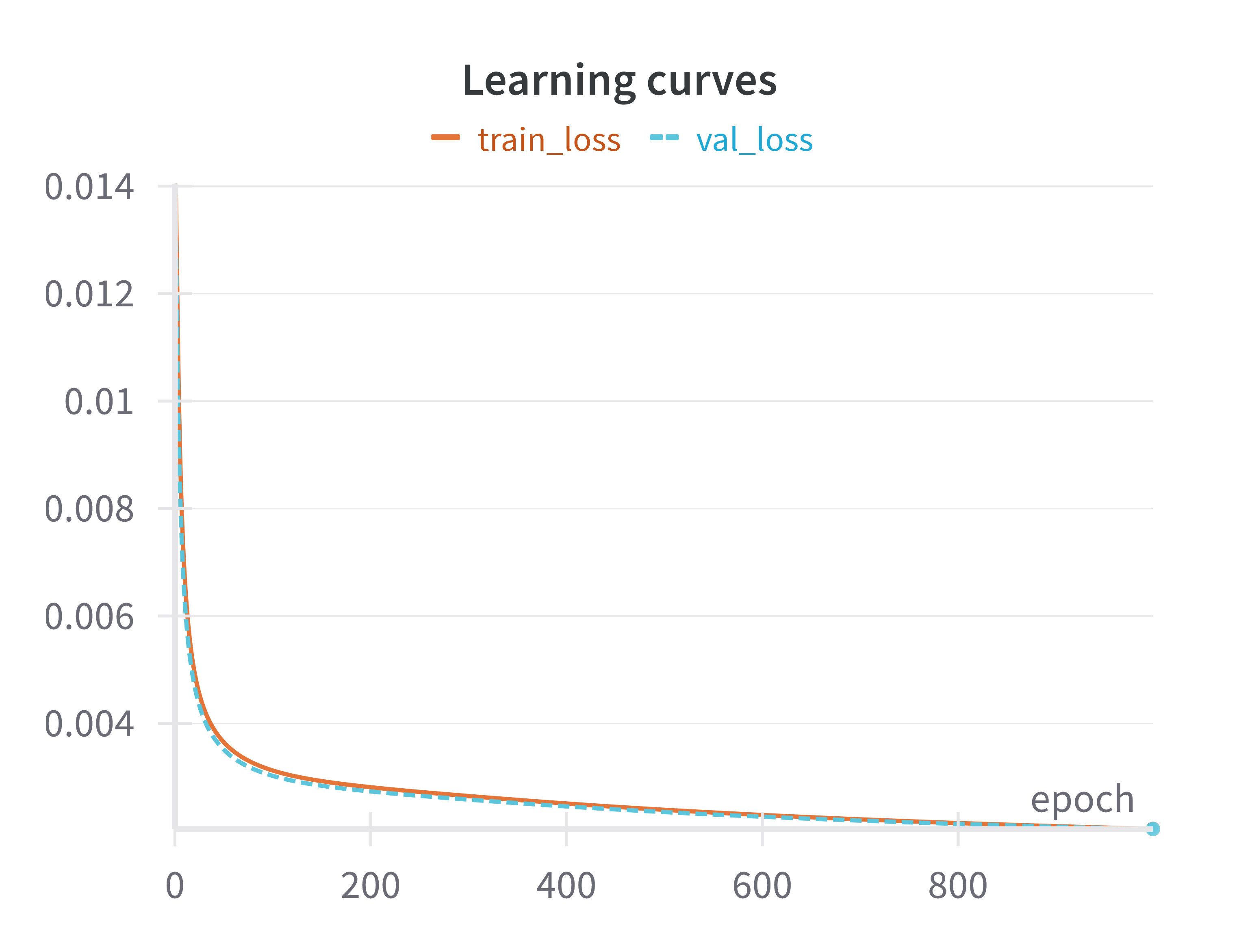}}
    \caption{Learning curves of Highway autoencoder training.}
    \label{fig:highway_ae_learning_curves}
\end{figure}

In the \ref{fig:highway_ae_learning_curves} are shown the learning curves obtained by the training of the best Highway autoencoder.

\section{Learning curves}

\begin{figure}[h]
    \centerline{\includegraphics[width=0.49\textwidth]{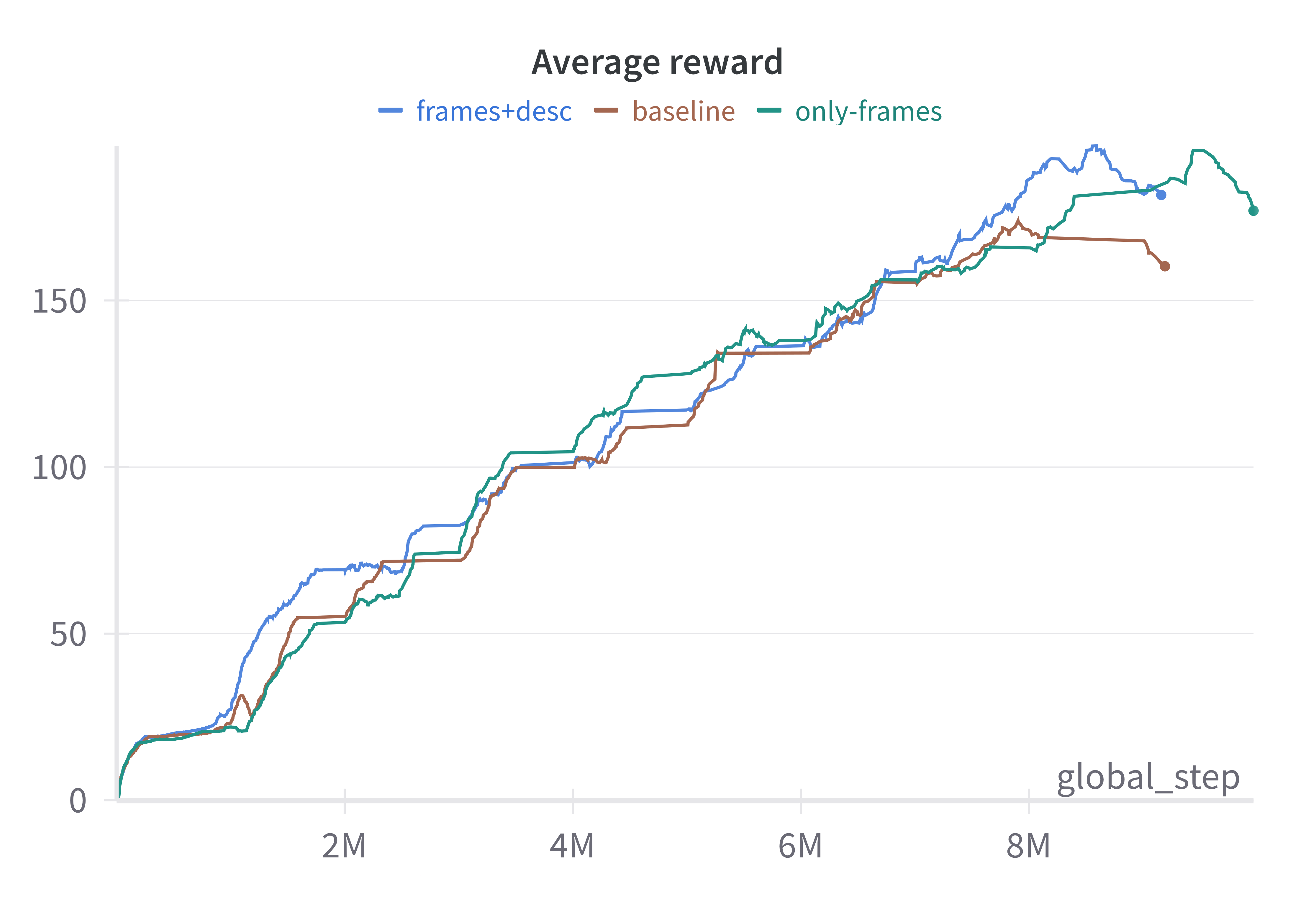}}
    \caption{Reward over time during FAST tests on Indiana task. The brown line corresponds to the baseline performance (SAC without transfer), the green line to the FAST performance using only frames, and the blue line to the FAST performance using frames and textual description.}
    \label{fig:best_transfer_indiana}
\end{figure}

\begin{figure}[h]
    \centerline{\includegraphics[width=0.49\textwidth]{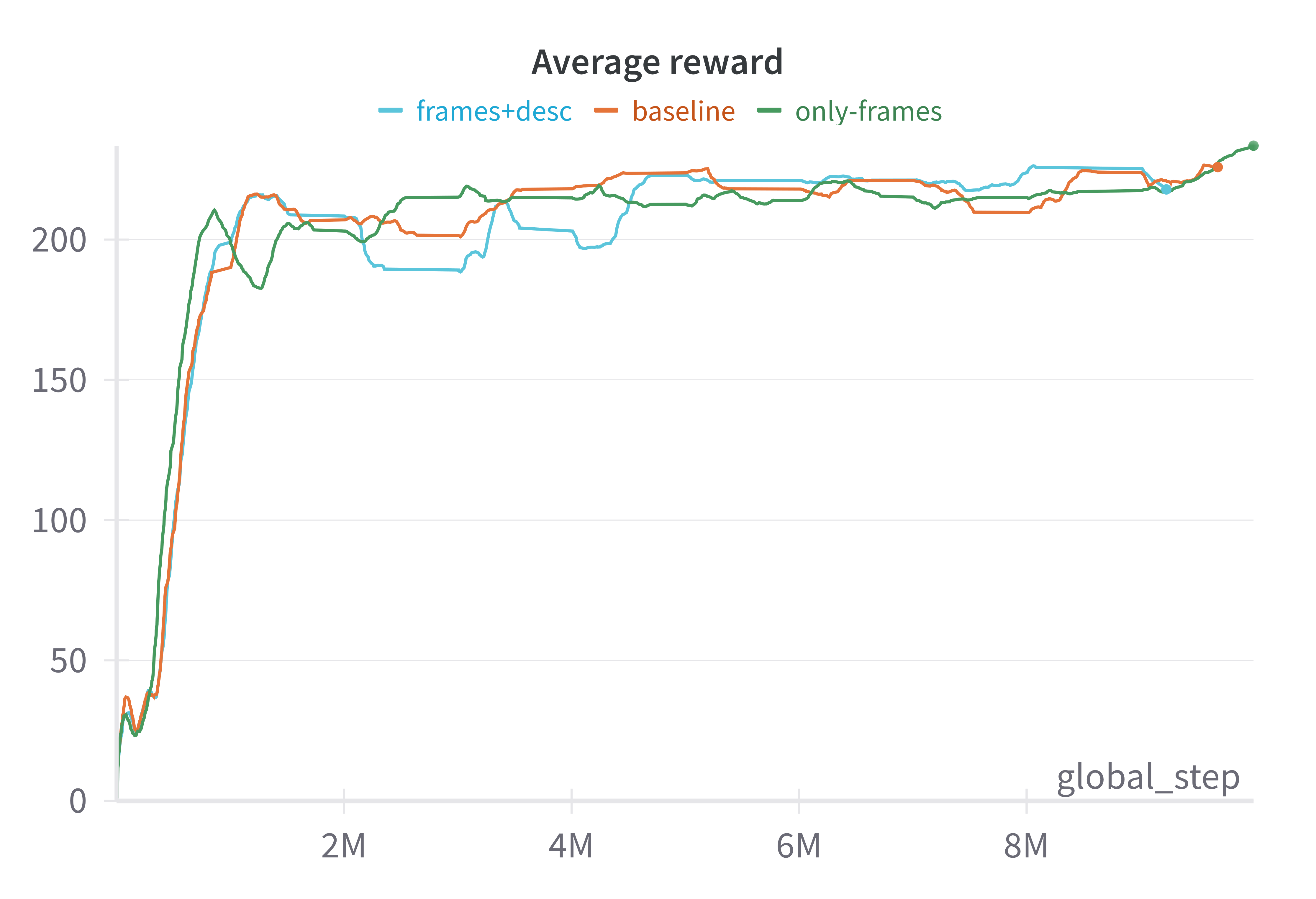}}
    \caption{Reward over time during FAST tests on Racetrack task. The brown line corresponds to the baseline performance (SAC without transfer), the green line to the FAST performance using only frames, and the cyan line to the FAST performance using frames and textual description.}
    \label{fig:best_transfer_racetrack}
\end{figure}

\begin{figure}[h]
    \centerline{\includegraphics[width=0.49\textwidth]{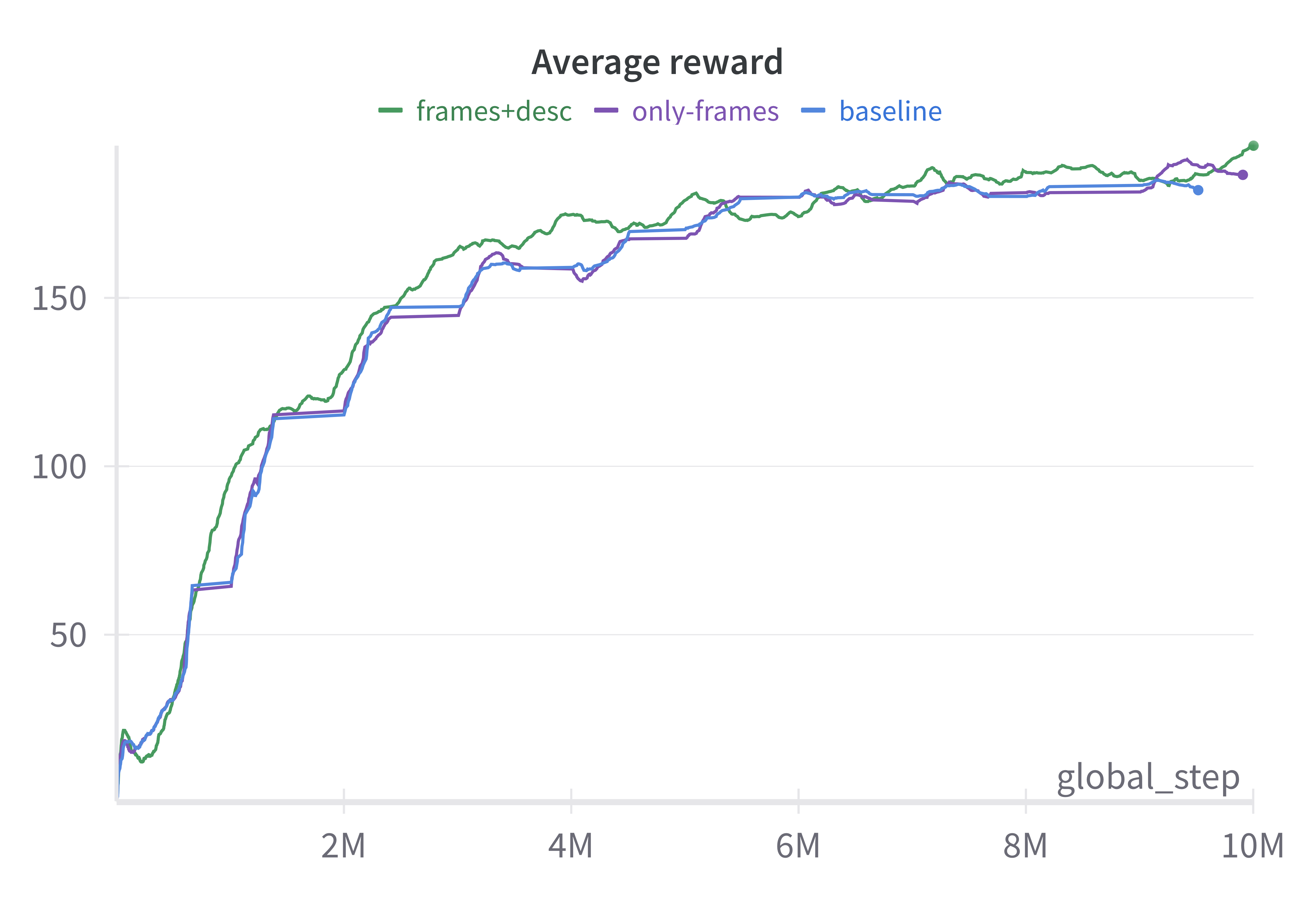}}
    \caption{Reward over time during FAST tests on Custom Racetrack task. The blue line corresponds to the baseline performance (SAC without transfer), the violet line to the FAST performance using only frames, and the green line to the FAST performance using frames and textual description.}
    \label{fig:best_transfer_custom_racetrack}
\end{figure}

In this section we report the learning curves of the target tasks considering all the variants. As shown in Fig. \ref{fig:best_transfer_indiana}, \ref{fig:best_transfer_racetrack}, and \ref{fig:best_transfer_custom_racetrack}, the FAST approach demonstrates slightly faster learning than the baseline while comparing only the reward score.

\section{Complex racetrack ablation study}
Given the extended length of the Custom Racetrack and agents did not complete a full lap in the initial experiments, we increased the maximum episode time from 150 to 400 seconds to evaluate the best-performing agent more thoroughly. Table \ref{tab:custom_racetrack_extended_time_transfer_metrics} reports the metrics from evaluations conducted on episodes with a maximum duration of 400 seconds. Performance generally improves as the maximum episode duration increases, with the exception of speed.

\begin{table}[!h]
    \caption{Metrics values of the best models on Custom Racetrack with extended time. \texttt{B} is the SAC baseline, \texttt{F} the proposed approach with frames only and \texttt{FT} with frames and textual descriptions.} 
    \begin{center}
        \begin{tabular}{c|c|c|c}
            & \textbf{B10} & \textbf{F10} & \textbf{FT10}\\
            \hline
            AD & $467.84_{\pm 213.31}$ & \boldmath{$515.46_{\pm 179.67}$} & $479.15_{\pm 301.46}$ \\[0.1cm]
            AR & $406.56_{\pm 217.03}$ & \boldmath{$438.29_{\pm 165.51}$} & $291.54_{\pm 190.55}$ \\[0.1cm]
            AL & $0.31_{\pm 0.34}$ & \boldmath{$0.35_{\pm 0.35}$} & $0.37_{\pm 0.44}$ \\[0.1cm]
            AS & $2.62_{\pm 3.88}$ & $2.39_{\pm 3.75}$ & \boldmath{$3.78_{\pm 4.90}$} \\[0.1cm]
        \end{tabular}
        \label{tab:custom_racetrack_extended_time_transfer_metrics}
    \end{center}
\end{table}

\pagebreak

\section{Full grid analysis}
As final experiment, a simultaneous tuning of all hyperparameters was conducted, encompassing both SAC and pipeline parameters. Previously, these sets were optimized independently. This comprehensive search aimed to determine whether a joint optimization could produce superior results.

\subsection{Setup}
Due to time constraints, a random search, rather than a grid search, was conducted on both variants exclusively in the \textit{Racetrack} environment. The pool of hyperparameter values is reported in Table \ref{tab:HypGridFull}.
\begin{table}[h]
    \caption{Values of hyper-parameters used to train SAC agent on the full set of hyperparameters.}
    \begin{center}
        \begin{tabular}{c|c}
            \textbf{Parameter} & \textbf{Values}\\
                     \hline
            learning rate & 1e-3, 1e-5, 1e-7 \\[0.1cm]
            batch size & 1024 \\[0.1cm]
            tau & 0.5, 0.6, 0.7 \\[0.1cm]
            gamma & 0.99 \\[0.1cm]
            gradient steps & 10 \\[0.1cm]
            k & 10, 100, 1000 \\[0.1cm]
            similarity threshold & 0.5, 0.6, 0.7 \\[0.1cm]
            total timesteps & 1e7 \\[0.1cm]
        \end{tabular}
        \label{tab:HypGridFull}
    \end{center}
\end{table}

\subsection{Results}
Table \ref{tab:BestHypGridFull} shows the best hyperparameters values found for each variant. When we compare these with the values found on Racetrack with separate training (Tab. \ref{tab:SAC_hp}-\ref{tab:transfer_hp_of}), we see similar hyperparameters across setups. One notable difference is the $\tau$ value, in the only frames full hyperparameters setting, it’s 0.6, whereas in the separate approach, it’s 0.5. Another difference is the similarity threshold value, in the separate approach is 0.5 for only frames flavor and 0.6 for frames+desc flavor, instead, in the full hyperparameters setting the values are inverted.

\begin{table}[h]
    \caption{Best values of hyper-parameters for the full train test in both flavors. \texttt{F} the proposed approach with frames only and \texttt{FT} with frames and textual descriptions.}
    \begin{center}
        \begin{tabular}{c|c|c}
        \textbf{Parameter} & \textbf{F} & \textbf{FT}\\
                 \hline
        learning rate & 1e-3 & 1e-3\\[0.1cm]
        tau & 0.6 & 0.5\\[0.1cm]
        k & 10 & 1000\\[0.1cm]
        similarity threshold & 0.6 & 0.5\\[0.1cm]
        \end{tabular}
        \label{tab:BestHypGridFull}
    \end{center}
\end{table}

Table \ref{tab:fullHP_racetrack_metrics} shows the evaluation metrics, comparing the current approach to the others. In terms of average distance (AD), the only frames approach with full hyperparameters set outperforms the rest, even though its average reward is lower than those from separate training setups. The only frames approach stands out as the best overall in both setups, with the highest average lap completions (AL) and speed (AS).

\begin{table}[t]
    \caption{Metrics values of the best models on Racetrack with full hyperparameters set. \texttt{F} the proposed approach with frames only, \texttt{FT} with frames and textual descriptions, \texttt{FF} with frames only and the full set of hyperparameters and \texttt{FFT} with frames and textual descriptions and the full set of hyperparameters.} 
    \begin{center}
        \begin{tabular}{c|c|c|c|c}
            & \textbf{AD} & \textbf{AR} & \textbf{AL} & \textbf{AS}\\
            \hline
            B & $818.64_{\pm 279.21}$ & $207.53_{\pm 70.10}$ & $2.23_{\pm 0.78}$ & $6.40_{\pm 1.10}$ \\[0.1cm]
            F & $999.91_{\pm 336.73}$ & \boldmath{$230.96_{\pm 68.28}$} & $2.75_{\pm 0.92}$ & $7.34_{\pm 1.17}$ \\[0.1cm]
            FT & $783.71_{\pm 228.82}$ & $229.80_{\pm 58.95}$ & $2.14_{\pm 0.64}$ & $5.74_{\pm 1.05}$ \\[0.1cm]
            FF & \boldmath{$1003.86_{\pm 306.67}$} & $223.21_{\pm 66.31}$ & \boldmath{$2.76_{\pm 0.86}$} & \boldmath{$7.67_{\pm 2.17}$} \\[0.1cm]
            FFT & $923.79_{\pm 245.93}$ & $229.08_{\pm 60.27}$ & $2.53_{\pm 0.68}$ & $6.73_{\pm 0.66}$ \\[0.1cm]
        \end{tabular}
        \label{tab:fullHP_racetrack_metrics}
    \end{center}
\end{table}

\end{document}